\documentclass{article}
\usepackage{graphicx}
\usepackage{bm}
\usepackage{array}
\usepackage{subcaption}
\usepackage[margin=1in]{geometry}
\usepackage{amsmath}
\usepackage{bbm}
\usepackage{multirow}
\usepackage{tabularx}
\usepackage{siunitx}
\usepackage{float}
\usepackage{natbib}
\usepackage{authblk}
\usepackage{url}

\title{Convolution Smoothed Quantile Regression for XGBoost }
\author[1]{Mandy Yao}
\author[1]{Meredith Franklin}
\affil[1]{Department of Statistical Sciences, University of Toronto}
\date{}

\begin{document}

\maketitle
\section{Abstract}
The increasing availability of large and complex datasets across many scientific disciplines has led to widespread adoption of machine learning (ML) for prediction. However, most ML algorithms focus on point estimation and provide limited information about predictive uncertainty or the conditional distribution of the response, restricting their ability to characterize rare or extreme outcomes. We develop QXGB, a quantile-based gradient boosting framework, and introduce a convolution smoothed loss within it that estimates conditional quantiles for constructing dense cumulative distribution functions (CDFs), exceedance probabilities, and tail behaviour relevant to extreme outcomes. This approach preserves the computational efficiency of extreme gradient boosting while restoring the Hessian information XGBoost relies on for tree splitting, in turn providing interpretable measures of extreme value and exceedance probability predictions. We derive the gradients and Hessians needed to integrate convolution smoothed quantile loss with different kernel specifications into XGBoost, and with simulated data, benchmark this approach against alternative smoothed quantile regression losses, the native quantile objective in the XGBoost Python package, and independent versus multi-output tree estimation. The practical relevance is illustrated in an application predicting fine particulate matter (PM$_{2.5}$) in northern California, including periods where levels were elevated due to wildfire smoke. Our results show that convolution smoothed QXGB, particularly when paired with multi-output trees, delivers accurate predictions with near-zero quantile crossing, well-calibrated CDF and exceedance probability estimates, and useful tail characterization for extreme values. Interval estimation is also evaluated as a measure of data spread.

\section{Introduction}

Many scientific applications are driven not by average behavior, but by rare but consequential extremes, such as wildfire smoke episodes, heat waves, and other threshold-crossing events. In these settings, point prediction alone is not sufficient; it is also important to predict how likely an outcome is to exceed a meaningful threshold, or how the upper tail of the predictive distribution behaves. This motivates methods that estimate conditional quantiles, exceedance probabilities, and other tail summaries directly, rather than relying only on point forecasts or mean-based uncertainty measures.

A natural way to address this need is through quantile regression, which directly estimates conditional quantiles without imposing a parametric form on the response distribution \citep{koenker1978}. This is especially appealing in settings with skewness, heavy tails, or rare extreme events, where mean-based models may obscure the behavior of the upper tail \citep{lakshminarayanan2017, 374138, pmlr-v48-gal16}. By estimating a dense sequence of these quantiles, the full conditional cumulative distribution function (CDF) can be reconstructed, and exceedance probabilities for extreme values can be derived to assess scientifically relevant thresholds \citep{franklin2026modeling}. In this way, quantile-based methods provide a flexible framework for characterizing both tail risk and predictive spread. However, existing quantile-boosting implementations can be difficult to optimize when many quantile levels are fit jointly, and their performance in the tail region may be uneven when the data are volatile or exhibit abrupt extremes \citep{velthoen2023}.

In this work, we focus on Extreme Gradient Boosting (XGBoost), a widely used ML method in both research and industry (\cite{Chen_Guestrin_2016}). To extend XGBoost from point prediction to distributional prediction, we adapt it for quantile regression and estimate a dense grid of conditional quantiles. However, quantile boosting introduces an important optimization challenge. The standard pinball loss is not differentiable everywhere, including at the zero residual, which can hinder convergence and lead to suboptimal parameter updates (\cite{horowitz1998, 2003Geop...68.1310G}). Furthermore, the second derivative is zero almost everywhere, which reduces curvature information and can slow optimization. 

Several methods have been proposed to address these issues. The Python xgboost package includes a direct quantile regression option, but it still optimizes the quantile (pinball) loss, and therefore does not remove the underlying non-smoothness of the objective  (\cite{Chen_Guestrin_2016}). Two alternative approaches introduce differentiable surrogates for the pinball loss: \cite{yin2023quantileextremegradientboosting} propose a Huber pinball loss that smooths the loss function to ensure continuous differentiability, while \cite{sluijterman2024} propose an arctan pinball loss that is not only continuously differentiable but also ensures a nonzero second derivative.

Building on these ideas, we integrate convolution-smoothed quantile regression \citep{tanwangzhou2021, He_Pan_Tan_Zhou_2023} into XGBoost, which smooths the pinball loss via kernel convolution while preserving the second-order gradient information XGBoost relies on for tree splitting. We compare this method against existing alternatives, including the Python package implementation and the aforementioned smoothed losses, in simulation studies designed to reflect volatile data with flat and extreme regions. We also compare estimating the dense quantile grid independently versus jointly with multi-output trees, and introduce a target standardization step that stabilizes the smoothed objectives against the large errors encountered at unmonitored spatial locations. Our focus is on dense cumulative distribution function (CDF) estimation and exceedance probability estimation for scientifically relevant thresholds. Finally, we illustrate the method in an application to wildfire-related PM$_{2.5}$ prediction in northern California, where recent work documents the role of smoke-driven extremes \citep{burke2023} and the use of machine learning to estimate air pollution exposures \citep{childs2022}. We assess performance using standard coverage and width diagnostics and proper interval scoring rules (\cite{gneiting2007}), and assess full probabilistic distributions using Continuous Ranked Probability Scores (CRPS) and quantile crossing rates.

\section{Methods}
Gradient boosting is an additive ensemble method that builds decision trees sequentially to minimize a specified loss function. Extreme Gradient Boosting (XGBoost) optimizes this process by supporting parallel processing and hardware acceleration, while adding a regularized objective function and second-order gradients (Hessians) to prevent overfitting and improve training efficiency. To extend this framework from point prediction to distributional prediction, we integrate quantile regression, which estimates conditional quantiles by minimizing the pinball loss function rather than the conditional mean. Detailed mathematical formulations of gradient boosting, XGBoost, and standard quantile regression are provided in the Appendix.

\subsection{Quantile Objective Functions}
By incorporating quantile regression into XGBoost, we can construct point-wise prediction intervals of specific quantiles. This is done simply by choosing the pinball loss function as the loss function that we minimize in the XGBoost algorithm. However, the pinball loss function is not differentiable when $y=\hat{y}$ and the second derivative is zero almost everywhere. This is an issue because XGBoost uses the hessian of the loss function during optimization. We consider three existing strategies to address this problem: the strategy used by the Python library xgboost, a Huber pinball loss function, and an arctan pinball loss function. We then introduce our proposed approach, convolution smoothed quantile regression.

\subsubsection{Python's xgboost}
Python's xgboost package has added an option to specify the objective function as "reg:quantileerror" to perform XGBoost with quantile regression, so that specific quantiles of the target variable's conditional distribution can be predicted (\cite{Chen_Guestrin_2016}). Since the gradient of the pinball loss function is not differentiable when $y=\hat{y}$, Python’s xgboost package uses a subgradient to approximate the gradient:

\[
\frac{\partial L_\tau(y, \hat{y})}{\partial \hat{y}} =
\begin{cases} 
-\tau & \text{if } y > \hat{y} \\
1-\tau & \text{if } y \leq \hat{y}
\end{cases}
\]

Since the Hessian is either zero or undefined, Python’s \texttt{xgboost} package sets it to the sample weights (equal to 1 by default). In this case, because the Hessian is constant across samples, there is no differentiation in the curvature contribution of individual samples beyond their weights. The calculation for the gain for a split is also simplified and therefore faster since the Hessian term is the same across all samples. However, because the Hessian is constant, the procedure is effectively first-order in spirit (i.e., gradient boosting), forfeiting some of the speed and efficiency advantages typical of second-order XGBoost updates.

\subsubsection{Huber Pinball Loss}
To ensure that the hessian of the loss function can be utilized in \texttt{xgboost}, we can use a smoothed approximation of the loss function whose second derivative is not 0 everywhere. \cite{yin2023quantileextremegradientboosting} approach this by using the Huber function, which is a smooth approximation of the absolute function 

\[
h_v(t) =
\begin{cases} 
\frac{t^2}{2v} & \text{if } |t|\leq v \\
|t|-\frac{v}{2} & \text{if } |t|>v,
\end{cases}
\]

where $v$ is a positive real number. By replacing $(y-\hat{y})$ with the huber function $h_v(t)$ in the pinball loss function, we get 

\[
\rho_\tau =
\begin{cases} 
(\tau-1)(|t|-\frac{v}{2}) & \text{if } t<-v \\
(\tau-1)(\frac{t^2}{2v}) & \text{if } -v\leq t<0 \\
\tau(\frac{t^2}{2v}) & \text{if } 0\leq t<v \\
\tau(|t|-\frac{v}{2}) & \text{if } t>v.
\end{cases}
\]

This function is continuously differentiable; its second derivative equals $1/v$ on $|t|<v$ and $0$ outside, so curvature is limited to a central band. It is important to choose $v$ carefully to obtain accurate estimates while still taking advantage of the use of Hessians in XGBoost. If the Huber loss function is too smooth, the quantiles will be biased, but if it is too similar to the original pinball loss function, it may behave similarly to the built-in Python xgboost function, which mostly reduces to first-order gradient boosting. 

\subsubsection{Arctan Pinball Loss}
Since the second derivative of the Huber pinball loss is zero outside the region $|t| \le v$,  training can resemble first-order gradient descent with a very low learning rate. To address this, \cite{sluijterman2024} introduced the arctan pinball loss function
\[L_{\tau,s}^{\text{(arctan)}}(y, \hat{y}) = 
\left(\tau - 0.5 + \frac{\arctan\left({\frac{y - \hat{y}}{s}}\right)}{\pi}\right)(y - \hat{y}) + \frac{s}{\pi}
\]
where $s>0$ is a smoothing parameter. This function ensures a nonzero second derivative everywhere. 

\subsubsection{Convolution Smoothing Approach}
\cite{He_Pan_Tan_Zhou_2023} devised a smoothing approach using convolution with kernel functions. The motivation came from another smoothing approach by \cite{horowitz1998}. If we let $u=y-\hat{y}$, the pinball loss function can be rewritten as 
\[\rho_\tau(u)=u\{\tau-\mathbbm{1}(u<0)\}.\]
Horowitz proposed to directly smooth this function to obtain
\[l_h^{Horo}(u)=u\{\tau-\mathcal{G}(-u/h)\},\]
where $\mathcal{G}(\cdot)$ is a smooth function that takes values between 0 and 1, and $h>0$ is a smoothing parameter, which we call a bandwidth. The issue with this approach is that the smoothness of the function causes it to lose its convexity, which results in optimization issues, especially in large dimensions.

If we let $Q(\bm\beta)=\mathbbm{E}\{\hat Q(\bm\beta)\}$ be the population pinball loss function, the first-order condition for convexity is satisfied for the population parameter $\bm\beta^*$:
\[\nabla Q(\bm{\beta}^*) = \mathbbm{E} \left[ \left( \mathbbm{1}\{ y < \mathbf{x}^\top \bm{\beta} \} - \tau \right) \mathbf{x} \right] \bigg|_{\bm{\beta} = \bm{\beta}^*} = \mathbf{0}.\]

\cite{He_Pan_Tan_Zhou_2023} used this to motivate a smoothed estimating equation (SEE) estimator, which is defined as the solution to the smoothed moment condition
\[\frac{1}{n} \sum_{i=1}^n \left[ \mathcal{G} \left( \frac{\langle \mathbf{x}_i, \bm{\beta} \rangle - y_i}{h} \right) - \tau \right] \mathbf{x}_i = \mathbf{0}.
\]

This SEE estimator can also be defined as a minimizer of an empirical smoothed loss function. Let $K(\cdot)$ be a kernel function that integrates to one, $h>0$ be a bandwidth, and let
\[
K_h(u) = h^{-1} K(u/h), \quad 
\mathcal{K}_h(u) = \mathcal{K}(u/h), \quad 
\text{and} \quad 
\mathcal{K}(u) = \int_{-\infty}^u K(v) \, dv, \quad u \in \mathbbm{R}.
\]

Then in the smoothing method proposed by \cite{He_Pan_Tan_Zhou_2023}, we minimize
\[
\widehat{Q}_h(\bm{\beta}) = \frac{1}{n} \sum_{i=1}^n \ell_h(y_i - \langle \mathbf{x}_i, \bm{\beta} \rangle) 
\quad \text{with} \quad 
\ell_h(u) = (\rho_\tau * K_h)(u) = \int_{-\infty}^\infty \rho_\tau(v) K_h(v - u) \, dv,
\]

where $*$ is the convolution operator. This empirical smoothed loss function is globally convex, in contrast to the smoothed loss function by Horowitz. In particular, if $K$ is non-negative, $\hat Q_h(\cdot)$ is a convex function for any $h>0$. The resulting estimator is given by
\[
\widehat{\bm{\beta}}_h = \widehat{\bm{\beta}}_h(\tau) \in \arg\min_{\bm{\beta} \in \mathbbm{R}^p} \widehat{Q}_h(\bm{\beta}).
\]
This estimator satisfies the first-order optimality condition  $\nabla Q_h(\hat{\bm{\beta}}_h)=\bm0$.

In addition, the smoothed function is twice continuously differentiable, with gradient and hessian matrices given by

\[
\nabla \widehat{Q}_h(\bm{\beta}) 
= \frac{1}{n} \sum_{i=1}^n \left\{ \mathcal{K}_h\left( \langle y_i-\mathbf{x}_i, \bm{\beta} \rangle \right) - \tau \right\} \mathbf{x}_i
\quad \text{and} \quad
\nabla^2 \widehat{Q}_h(\bm{\beta}) 
= \frac{1}{n} \sum_{i=1}^n K_h\left( y_i - \langle \mathbf{x}_i, \bm{\beta} \rangle \right) \mathbf{x}_i \mathbf{x}_i^\top.
\]

\subsubsection{Convolutional Smoothing Computations for Commonly Used Kernels}
In remark 3.1 of \cite{He_Pan_Tan_Zhou_2023}, the loss function was rewritten as
\[\rho_\tau(u) = \frac{|u|}{2} + \left(\tau - \frac{1}{2}\right)u.\]

After convolution with a kernel, we get that
\[\ell_h(u) = \frac{1}{2} \int_{-\infty}^\infty |u + hv| K(v) \, dv + \left(\tau - \frac{1}{2}\right)u.
\]

It can then be shown that 
\[\ell_h'(u)=\tau-\mathcal{K}_h(u),\quad \ell_h''(u)= \frac{1}{h} K\left( \frac{u}{h} \right)\]

To demonstrate this method in practice, we compute the convolution smoothed pinball loss functions, as well as their gradients and Hessians, for two kernels. We write down the kernels $K(u)$ and their convolution-smoothed pinball loss functions $\ell_h(u)$, along with the gradients and Hessians $\ell_h'(u)$ and $\ell_h''(u)$. Another four commonly used kernels are provided in the Appendix.

\begin{enumerate}
  \item Gaussian kernel \(K(u) = (2\pi)^{-1/2}e^{-u^2/2}\):
  \[
    \ell_h(u) = \frac{h}{2}\,\ell^G(u/h) + \Bigl(\tau - \tfrac{1}{2}\Bigr)u,
    \quad \ell^G(u) := \sqrt{\tfrac{2}{\pi}}e^{-u^2/2} + u\{1 - 2\Phi(-u)\}.
  \]
  \[
\ell_h'(u) = \tau - \Phi\left( -\frac{u}{h} \right), \quad
\ell_h''(u) = \frac{1}{h\sqrt{2\pi}} e^{-u^2 / (2h^2)}
\]
    \item Laplace Kernel \(K(u) = \frac{1}{2} e^{-|u|}\):

\[
    \ell_h(u) = \frac{h}{2}\,\ell^{La}(u/h) + \Bigl(\tau - \tfrac{1}{2}\Bigr)u,
    \quad \ell^{La}(u) := \frac{1}{2h}\int^\infty_{-\infty}|u+hv|\cdot e^{-|v|}dv.
  \]
\[
\ell_h'(u) =
\begin{cases}
\tau - \frac{1}{2} e^{u/h}, & u < 0 \\
\tau - 1 + \frac{1}{2} e^{-u/h}, & u \geq 0
\end{cases},
\quad
\ell_h''(u) = \frac{1}{2h} e^{-|u|/h}
\]
\end{enumerate}

\subsection{Independent versus Multi-Output Trees}
\label{sec:multi-output-trees}
The quantile objective functions described above can be incorporated into XGBoost in two different ways. The default approach fits a separate boosted ensemble for each quantile level $\tau$ using independent additive trees and the exact greedy split algorithm, so that each quantile's tree structure is learned entirely independently of the others. Because the resulting quantile functions are estimated separately, there is no mechanism enforcing that $\hat{y}_{\tau_i} \le \hat{y}_{\tau_j}$ for $\tau_i < \tau_j$, and quantile crossing frequently results, particularly for out-of-distribution feature values.

Alternatively, XGBoost's histogram-based tree method supports jointly estimating multiple outputs within a single boosted ensemble using multi-output trees, where each split is chosen to minimize a loss aggregated across quantile levels and every leaf stores a vector of predicted quantiles rather than a scalar. Because a single tree structure is shared across quantile levels, the fitted values for different quantiles are coupled rather than independent, allowing information to be shared across quantiles and substantially reducing the incidence of quantile crossing (\cite{sluijterman2024}). We evaluate both the independent and multi-output tree approaches for each of the quantile objective functions described above.

\subsection{Dense CDF Estimation and Target Standardization}
\label{sec:target-standardization}
Extreme value characterization is one important application of conditional quantile regression, but estimating a dense conditional cumulative distribution function (CDF) is useful more broadly for uncertainty quantification, probabilistic prediction, and risk assessment (\cite{sluijterman2024}). Rather than estimating only a small number of quantiles (e.g. median and upper/lower prediction bounds), we extend the smoothed quantile framework to estimate a dense grid of conditional quantiles 
\[\tau \in T = \{0.05, 0.15, 0.25, 0.35, 0.45, 0.55, 0.65, 0.75, 0.85, 0.95\},\]
which provides a discrete approximation of the conditional inverse CDF, $Q_Y(\tau\mid X)=F^{-1}_{Y\mid X}(\tau)$.

In applications such as spatiotemporal air quality modeling, predicting the precise point estimate of a pollutant is often less critical than knowing if it will cross a regulatory or health-based threshold. For example the US Environmental Protection Agency (EPA) sets a 24-hour ambient air quality standard for $PM_{2.5}$ at $35.0~\mu g/m^{3}$. From the dense grid, the aim is to accurately model the conditional exceedance probability, allowing decision-makers to assess the precise likelihood of a hazardous event:
\[P(Y>y_{thresh}|X)=1-F_{Y|X}(y_{thresh})\]
where in this example, $y_{thresh}=35.0~\mu g/m^{3}$. To do so, we sort the predicted quantiles for each observation to enforce monotonicity in the rare cases where crossing occurs, locate the two (now-ordered) predicted quantiles that bracket the 35.0 threshold, and use linear interpolation to extract the exceedance probability. This sorting step is applied only for the exceedance probability calculation; the unadjusted crossing rate metric introduced below is reported separately as a diagnostic of each method's intrinsic stability.

When estimating dense quantiles, large prediction errors can cause mathematical instability in the objective functions. Specifically, convolution-smoothed objectives (Laplace and Gaussian) rely on exponential functions ($e^z$ and $e^{-z}$) to compute their gradients and Hessians. If an unscaled error term $z$ becomes exceptionally large, the exponential function causes floating-point overflow, resulting in substantial quantile crossing (where $\hat{y}_{\tau_i} > \hat{y}_{\tau_j}$ for $\tau_i < \tau_j$).

To ensure mathematical stability against such large errors, we introduce a target standardization (Z-Scaling) step prior to tree fitting. We compute the mean $\mu_{train}$ and standard deviation $\sigma_{train}$ of the training target variable, and scale the targets as $y_{scaled} = (y - \mu_{train}) / \sigma_{train}$. This standardization strictly bounds the magnitude of the errors fed into the exponential kernels. Once the XGBoost trees output the scaled quantile predictions, they are inverse-transformed back to the original domain: $\hat{y}_{real} = (\hat{y}_{scaled} \times \sigma_{train}) + \mu_{train}$.

\subsection{Metrics for Model Predictions}
To evaluate both estimated cumulative distribution functions and exceedance probabilities constructed from our XGBoost-based quantile regression models, we utilize a suite of evaluation metrics.

To evaluate estimated CDFs constructed from the model predictions for a dense grid of quantiles, we use the Continuous Ranked Probability Score (CRPS) \citep{hersbach2000, gneiting2007}. The CRPS is a strictly proper scoring rule that assesses both the calibration and sharpness of a predictive distribution. It is defined as the integrated squared difference between the predicted cumulative distribution function $F$ and the true observation step-function:
\begin{equation}
    \text{CRPS}(F, y) = \int_{-\infty}^\infty \left( F(x) - \mathbbm{1}\{x \geq y\} \right)^2 dx
\end{equation}
In the context of quantile regression, this integral can be equivalently expressed as the integral of the pinball loss $\rho_\tau$ across all quantile levels $\tau \in [0, 1]$:
\begin{equation}
    \text{CRPS}(F, y) = 2 \int_0^1 \rho_\tau(y, \hat{y}_\tau) d\tau
\end{equation}
Because we evaluate a discrete grid of quantiles $\mathcal{T} = \{\tau_1, \dots, \tau_K\}$, we approximate this continuous integral using the trapezoidal rule:
\begin{equation}
    \widehat{\text{CRPS}} = \sum_{k=1}^{K-1} (\tau_{k+1} - \tau_k) \left( \rho_{\tau_k}(y, \hat{y}_{\tau_k}) + \rho_{\tau_{k+1}}(y, \hat{y}_{\tau_{k+1}}) \right)
\end{equation}
A lower CRPS indicates a forecast that is both sharp and well-calibrated.

A fundamental property of a valid cumulative distribution function is monotonicity. When quantile models are trained independently rather than jointly with multi-output trees (Section~\ref{sec:multi-output-trees}), extreme out-of-distribution feature spaces can cause the estimated quantile functions to cross, rendering the resulting CDF invalid. For a sorted set of $K$ quantiles $\tau_1 < \tau_2 < \dots < \tau_K$, we define the crossing rate over $n$ samples as the empirical probability of a monotonicity violation:
\begin{equation}
    \text{Crossing Rate} = \frac{1}{n} \sum_{i=1}^n \mathbbm{1} \left\{ \exists k \in \{1, \dots, K-1\} \text{ s.t. } \hat{y}_{i, \tau_k} > \hat{y}_{i, \tau_{k+1}} \right\}
\end{equation}
where $\mathbbm{1}\{\cdot\}$ is the indicator function. This metric reflects both the gradient stability of the underlying loss function, since unstable Hessians affect tree splits and produce elevated crossing rates, and whether quantiles were estimated independently or jointly with multi-output trees.

To evaluate our ability to accurately map exceedance risk over specific thresholds, we utilize calibration bias, Brier skill score and ROC-AUC score.

The total number of observed exceedances, $N_{\text{obs}}$, is simply the sum of the true binary outcomes over the dataset:
\begin{equation}
    N_{\text{obs}} = \sum_{i=1}^N o_i
\end{equation}
The expected number of predicted exceedances, $N_{\text{pred}}$, is computed as the sum of the predicted probabilities across all observations:
\begin{equation}
    N_{\text{pred}} = \sum_{i=1}^N \hat{p}_i
\end{equation}
Using these two quantities, we define calibration bias as the relative difference between predicted and observed exceedance counts:
\begin{equation}
    \text{Calibration Bias} = \frac{N_{\text{pred}} - N_{\text{obs}}}{N_{\text{obs}}}
\end{equation}
A calibration bias near zero indicates that a model's predicted exceedance probabilities are neither systematically over- nor under-confident; positive values indicate overprediction of risk and negative values indicate underprediction.

The Brier Score \citep{brier1950} is a strictly proper scoring rule used to evaluate the accuracy of probabilistic forecasts for binary events. It is defined as the mean squared error between the predicted probabilities and the actual binary outcomes:
\begin{equation}
    BS = \frac{1}{N} \sum_{i=1}^N (\hat{p}_i - o_i)^2
\end{equation}
The Brier Score ranges from 0 to 1, with lower scores indicating greater accuracy. A score of 0 represents a perfect deterministic forecast.

To contextualize the Brier Score, the Brier Skill Score (BSS) \citep{murphy1973} compares the model's performance against a reference forecast ($BS_{\text{ref}}$). The standard reference is a static, base-rate climatology forecast, which simply assigns the empirical dataset exceedance rate, $\bar{o} = \frac{1}{N} \sum_{i=1}^N o_i$, to every prediction. The reference Brier Score is thus $BS_{\text{ref}} = \frac{1}{N} \sum_{i=1}^N (\bar{o} - o_i)^2$. 

The BSS is calculated as the relative improvement:
\begin{equation}
    BSS = 1 - \frac{BS_{\text{model}}}{BS_{\text{ref}}}
\end{equation}
The BSS provides an intuitive interpretation of model quality:
\begin{itemize}
    \item $BSS = 1$: A perfect probabilistic forecast.
    \item $BSS > 0$: The model provides an improvement over the baseline.
    \item $BSS = 0$: The model offers no improvement over simply predicting the constant baseline probability.
    \item $BSS < 0$: The model exhibits worse performance than the baseline forecast.
\end{itemize}

The Area Under the Receiver Operating Characteristic Curve (ROC AUC) measures the model's ability to successfully discriminate between exceedance and non-exceedance events. It plots the True Positive Rate (Sensitivity) against the False Positive Rate (1 - Specificity) across all possible probability decision thresholds. 

An AUC of 1.0 indicates perfect separation, while an AUC of 0.5 indicates performance equivalent to random guessing. While the Brier Score heavily penalizes poor absolute calibration, ROC AUC evaluates relative ranking.

\subsection{Interval Estimation}
In addition to the dense quantile grid used for CDF and exceedance probability estimation, a single pair of symmetric quantile estimates can be combined to form a prediction interval, a standard and widely used measure of predictive spread \citep{gneiting2014, pearce2018}. A prediction interval $[\hat{y}_{\alpha/2}, \hat{y}_{1-\alpha/2}]$, formed from quantile estimates at levels $\alpha/2$ and $1-\alpha/2$, has a nominal coverage probability of $(1-\alpha)\cdot 100\%$. Throughout this paper, we evaluate 90\% prediction intervals formed from the 0.05 and 0.95 quantile estimates, using each of the quantile objective functions and tree strategies described above.

To evaluate these intervals, we use three standard metrics: the prediction interval coverage probability (PICP), the prediction interval normalized average width (PINAW), and an interval scoring rule. These are widely used in practice because there is no single gold standard metric for interval evaluation \citep{Kabir_Khosravi_Hosen_Nahavandi_2018, Nourani_Paknezhad_Tanaka_2021a, pearce2018, Lai_Shi_Han_Shao_Qi_Li_2022}.

The PICP shows the proportion of the data that is within the target prediction interval created from estimating a lower and upper quantile separately. It is given by
\[\text{PICP}=\frac{1}{n}\sum^n_{i=1}c_i,\quad c_i=
\begin{cases}
    1 & t_i\in [\underline{y}_i,\overline{y}_i] \\
    0 \quad\text{otherwise}
\end{cases}\]
where n is the number of samples, $t_i$ is the true target value of the $i_{th}$ sample, $\underline{y}_i$ is the lower quantile estimate of the $i_{th}$ sample, and $\overline{y}_i$ is the upper quantile estimate of the $i_{th}$ sample. A higher PICP is more desirable since it indicates that the intervals produced by the quantile estimates cover more of the data on which we are estimating quantiles.

The PINAW quantifies the average interval width, normalized to make results comparable across different datasets. In this way, the metric can be compared across datasets. It is given by
\[\text{PINAW}=\frac{1}{R\times n}\sum^n_{i=1}(\overline{y}_i-\underline{y}_i), \qquad
R=\max_{1\le i\le n} y_i - \min_{1\le i\le n} y_i,
\]
where $R$ is the range of the observed data. A smaller PINAW is more desirable since intervals with large widths are not useful, as they do not provide much information about the target values.

As another metric to evaluate intervals, we use an interval scoring rule, which accounts for both the width and the coverage of the interval. It is a proper scoring rule, which means that the expected score is minimized if the predicted interval corresponds to the true quantiles of the underlying distribution at levels $\alpha/2$ and $1 - \alpha/2$ \citep{gneiting2007}. This means that a lower mean interval score across the test set indicates a superior model that provides well-calibrated and informative spread estimates. For a central $(1 - \alpha) \cdot 100\%$ interval defined by the lower bound $L$ and upper bound $U$, and given an observed value $y$, the interval score $S(L, U; y)$ is defined as:

\begin{equation}
    S(L, U; y) = (U - L) + \frac{2}{\alpha}(L - y)\mathbbm{1}\{y < L\} + \frac{2}{\alpha}(y - U)\mathbbm{1}\{y > U\}
\end{equation}

where $\mathbbm{1}\{\cdot\}$ denotes the indicator function. The first term, $(U - L)$, represents the width of the interval, and rewards narrower predictions. The next two terms add a penalty proportional to the distance of the observation $y$ from the nearest bound if $y$ falls outside the interval. When tuning hyperparameters for interval bounding, optimal parameters were selected to minimize the interval score, with a numerical penalty incorporated to heavily penalize quantile crossings during tuning.

\section{Simulation Study}
\begin{figure}[H]
\begin{center}
\includegraphics[scale=0.8]{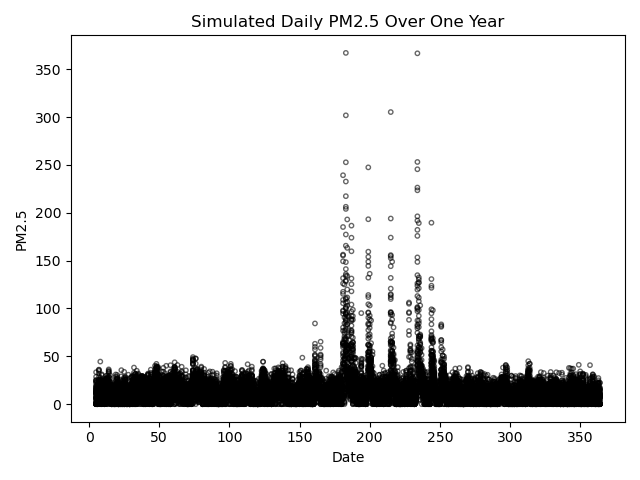}
\caption{Simulated daily PM$_{2.5}$ concentrations ($\mu$ g/m$^3$) for one year, with more extreme levels during the wildfire season.}
  \label{fig:sim-overview}
\end{center}
\label{fig:sim-data}
\end{figure}

We compare the performance of all previously described methods with the proposed convolution smoothed quantile regression using XGBoost via a simulation study. The simulated data were designed to capture data with significant volatility, and to mimic the spatiotemporal application described in the next section: multi-year, daily PM$_{2.5}$ measured at $S=50$ monitoring sites, with a pronounced elevation during the 2018 wildfire season (June--September; see Figure~\ref{fig:sim-overview}).

\subsection{Data Generating Process}

The goal of the data generating process was to create a spatio-temporal dataset across 50 randomly distributed geographical locations. To achieve this, we used a baseline Auto-Regressive (AR) process with correlated noise, supplemented by a conditional, spatially decaying shock mechanism for extreme events. 

First, we simulated the 50 geographical locations by randomly assigning coordinates in a 10 by 10 Euclidean space. We use an exponential covariance function $\Sigma(d)=\exp(-d/\phi),$ where $d$ is the Euclidean distance and $\phi$ is a spatial range parameter. We then use a first-order Auto-Regression (AR(1)) process to model the target variable, where the error term is spatially correlated Gaussian noise, to ensure that both temporal and spatial components are accounted for. 

In addition to this process, we also have a process to simulate extreme events. We do this by simulating 15 covariates, where the first 3 are used as "drivers" to trigger extreme values in the target variable, specifically in the fire season (days 152 to 273 of each year). Specifically, a risk score was calculated as a weighted linear combination of the 3 "driver" covariates, and an extreme event occurred with a probability of 10\% during the fire season, and with a probability of 5\% during the rest of the year. Then, if a spike occurred, we assign a fire epicenter based on the risk score map. The magnitude of the extreme values was scaled by the epicenter's risk value and spread to neighbouring locations using the function $\exp(-d/\phi)$. This helps to simulate a localized, high-magnitude event, which is capped at 500 during the fire season and 100 during the off-season. We also ensure that the final value is non-negative by replacing the value with 0 for negative values.

\subsection{Model Tuning and Evaluation}

For each simulated dataset, we tuned tree-level hyperparameters using an XGBoost model with squared-error loss (RMSE objective). We randomly split the data into training and testing sets, where 20\% of the data is used as the testing set and the remaining 80\% of the data is used as the training set. We then randomly sampled 100 hyperparameter combinations from a hyperparameter grid and ran 10-fold cross-validation (CV) on the training data to choose the best hyperparameters that produce the lowest RMSE. We then fixed the selected values across all quantile methods for comparability.

To rigorously tune the method-specific smoothing parameters, we introduced a spatial 5-fold cross-validation approach. The 50 simulated locations were randomly split into 5 spatial folds. We evaluated smoothing parameters from predefined grids ($t, s, h \in \{0.10, 0.25, 0.50, 0.75, 1.0\}$) and for the exceedance probability estimation, we selected the parameters that minimized the BSS. To avoid infinite value interval scores while penalizing quantile crossings, we incorporated a numerical penalty into the interval score calculation for rows where the estimated lower bound exceeded the upper bound.

Finally, to rigorously test the models' capacity for spatial prediction, we used a Leave-One-Site-Out Cross-Validation (LOSO-CV) framework. In this setup, the test set consists entirely of data from a single held-out location, while the training set comprises all historical data from the remaining 49 locations. We evaluated two distinct approaches for the tree method in Python's \texttt{xgboost} algorithm: independent additive trees using the exact greedy algorithm, and joint multi-output trees using the histogram tree method. Target standardization (Z-scaling) was applied to stabilize the convolution kernels against large deviations in spatial variance. 

\subsection{Comparing Convolution Smoothing Kernels}
For the convolution smoothing method, we compared the Laplace and Gaussian Kernels. The Laplace kernel, being heavier tailed, is expected to perform well for data with many outliers (e.g. the 2018 spike), whereas the Gaussian kernel is more standard and provides a smoother baseline for milder tails.

\subsection{Simulation Results for CDF Estimation}
Table \ref{tab:sim-crps-results2} summarizes the performance using the multi-output tree method for the dense CDF evaluation, while Table \ref{tab:sim-crps-results2-ind} in the Appendix shows the independent tree method. When predicting a dense grid of quantiles independently, the models suffer from substantial quantile crossing (e.g., average crossing rates up to 0.69 for the Huber pinball method and 0.43 for the convolution methods at lower smoothing parameters). However, estimating the quantiles jointly with multi-output trees effectively resolves this issue, with average crossing rates approaching 0.00 across all smoothed methods. This is likely due to the fact that we are using the same model for multiple quantiles, so that information across quantiles can be shared. The default Python method still suffers from a relatively high average crossing rate of 0.12. When tuning the smoothing parameters, we notice that there is a tradeoff between average CRPS and average crossing rates. Lower smoothing parameter values yield a model closer to the default Python method with lower average CRPS, while higher values yield a lower average crossing rate. This effect is most pronounced in the arctan pinball smoothing method, where the average CRPS increases substantially as the smoothing parameter increases. The Huber pinball smoothing method achieves substantially lower crossing rates as the smoothing parameter increases. The convolution methods achieve relatively stable and low CRPS values and crossing rates.

\begin{table}[H]
\caption{Average CRPS and Crossing Rates across all 50 simulated locations.}
\centering
\renewcommand{\arraystretch}{1.3}
\begin{tabularx}{\textwidth}{l >{\centering\arraybackslash}X *{2}{>{\centering\arraybackslash}X}}
\hline
\textbf{Method} & \textbf{Smoothing Parameter} & \textbf{Average CRPS} & \textbf{Average Crossing Rate} \\
\hline
Default Python         & \multirow{4}{*}{0.1} & 2.72 & 0.12 \\
Huber Pinball         &                      & 2.77 & 0.15 \\
Arctan Pinball         &                      & 2.74 & 0.01 \\
Convolution (Laplace)  &                      & 2.74 & 0.02 \\
Convolution (Gaussian) &                      & 2.75 & 0.03 \\
\hline
Default Python         & \multirow{4}{*}{0.5} & 2.72    & 0.12\\
Huber Pinball          &                      & 2.80 & 0.02 \\
Arctan Pinball         &                      & 3.24 & 0.00 \\
Convolution (Laplace)  &                      & 2.85 & 0.00 \\
Convolution (Gaussian) &                      & 2.79 & 0.00 \\
\hline
Default Python        & \multirow{4}{*}{1.0} & 2.72 & 0.12 \\
Huber Pinball         &                      & 2.83 & 0.01 \\
Arctan Pinball         &                      & 4.35 & 0.00 \\
Convolution (Laplace)  &                      & 3.36 & 0.00 \\
Convolution (Gaussian) &                      & 3.19 & 0.00 \\
\hline
\end{tabularx}
\label{tab:sim-crps-results2}
\end{table}

\subsection{Simulation Results for Exceedance Probabilities}
The exceedance probability results validate our ability to identify when values pass the $35.0~\mu g/m^{3}$ threshold. As shown in Table \ref{tab:exceedance-results-sim}, direct binary classification (Logistic baseline) overpredicts risk and has a negative BSS score. The Arctan method also overpredicts risk compared to other smoothing methods. The convolution smoothing methods have the largest BSS and provide the most accurate exceedance probability estimates compared to the other methods.

\begin{table}[H]
\caption{Exceedance Probability Results.}
\centering
\renewcommand{\arraystretch}{1.3}
\begin{tabularx}{\textwidth}{>{\raggedright\arraybackslash}X c c c c c}
\hline
\textbf{Model}                  & \textbf{Tuned Smoothing Parameter} & $\bm{N_{obs}}$ & $\bm{N_{pred}}$ & \textbf{BSS} & \textbf{ROC-AUC} \\
\hline
Logistic (Binary Baseline)      &       & 845 & 3962.50 & -0.87 & 0.80 \\
Method 1: QXGB (Default)        &       & 845 & 919.00  & 0.30 & 0.82 \\
Method 2: QXGB (Huber)          & t=0.1 & 845 & 806.50  & 0.30 & 0.81 \\
Method 3: QXGB (Arctan)         & s=0.05 & 845 & 1060.6  & 0.30 & 0.82 \\
Method 4: QXGB (Gaussian Tuned) & h=0.2 & 845 & 890.30  & 0.31 & 0.81 \\
Method 5: QXGB (Laplace Tuned)  & h=0.1 & 845 & 850.80  & 0.31 & 0.81 \\
\hline
\end{tabularx}
\label{tab:exceedance-results-sim}
\end{table}

Additional results for the exceedance probabilities are included in the appendix in Table \ref{tab:exceedance-results-sim50}, where the threshold was increased to $50.0~\mu g/m^{3}$. In this case, the results were worse for almost all methods in almost all metrics. In particular, the risk was more overestimated across most methods. This may be due to the new threshold representing a much rarer and far-right tail event in the simulated distribution. Since the data is more sparse in the far right tail of the distribution, it is more difficult to obtain accurate estimates of probabilities for extreme values.

\subsection{Simulation Results for Interval Estimation}
We fit the methods to estimate the 0.05 and 0.95 conditional quantiles and formed 90\% intervals from these estimates. We report PICP, PINAW, and interval score for each method. The interval results for predicting the 90\% prediction intervals using the joint multi-output tree method are shown in Table \ref{tab:sim-results}, while the independent tree results are located in the appendix in Table \ref{tab:sim-results2}.

The arctan pinball, convolution, and Python xgboost methods strongly outperformed the Huber pinball smoothing method. Achieving PICPs $\ge$ 0.9 was straightforward for arctan and convolution smoothing, but not attainable for Huber pinball. Laplace and Gaussian kernels produced similar results for the convolution smoothing method and slightly outperformed the arctan method by having lower PINAW values. The Python xgboost method also often produces intervals with coverage less than 0.9, with no smoothing parameter to adjust the coverage. However, unlike Huber pinball, its narrow intervals are not offset by severe undercoverage, and it achieves the lowest interval score of all methods; Huber pinball has a marginally lower PINAW, but its poor coverage (PICP = 0.73) results in the highest interval score.

\begin{table}[H]
\caption{PICP, PINAW, and interval score for 90\% intervals formed from the 0.05 and 0.95 quantiles}
\centering
\renewcommand{\arraystretch}{1.3}
\begin{tabularx}{\textwidth}{p{1.7cm} *{5}{>{\centering\arraybackslash}X}}
\hline
\textbf{Simulated Data} & \textbf{Python} &\textbf{Huber Pinball} & \textbf{Arctan Pinball} & \textbf{Convolution (Laplace)} & \textbf{Convolution (Gaussian)} \\
\hline
\multicolumn{6}{>{\raggedright\arraybackslash}p{\linewidth}}{\textbf{With Spike}}\\
\textit{PICP} & 0.86  & 0.73 & 0.97 & 0.91 & 0.94 \\
\textit{PINAW} & 0.18  & 0.14 & 0.25 & 0.19 & 0.20 \\
\textit{Interval Score} & 49.48 & 55.71 & 53.86 & 50.26 & 51.20 \\
\textit{Tuned Smoothing Parameter} &  & t=0.5 & s=0.1 & h=0.1 & h=0.25 \\
\hline
\end{tabularx}
\label{tab:sim-results}
\end{table}

A sensitivity analysis was done for the interval study to assess how sensitive the results are to changes in the smoothing parameters. These results are shown in detail in Table \ref{tab:sim-interval-sensitivity} in the appendix. In general, the arctan pinball method is much more sensitive to changes in the smoothing parameters than the other methods, with PINAW increasing from 0.25 to 1.43 as the smoothing parameter increases from 0.1 to 1.0, creating intervals too wide to be useful.

\section{Application to PM$_{2.5}$ Prediction}
\subsection{Data}
\subsubsection{PM$_{2.5}$ Measurements}

\begin{figure}[h!]
 \centering
  \begin{subfigure}[t]{0.45\textwidth}
    \centering
    \includegraphics[width=\linewidth]{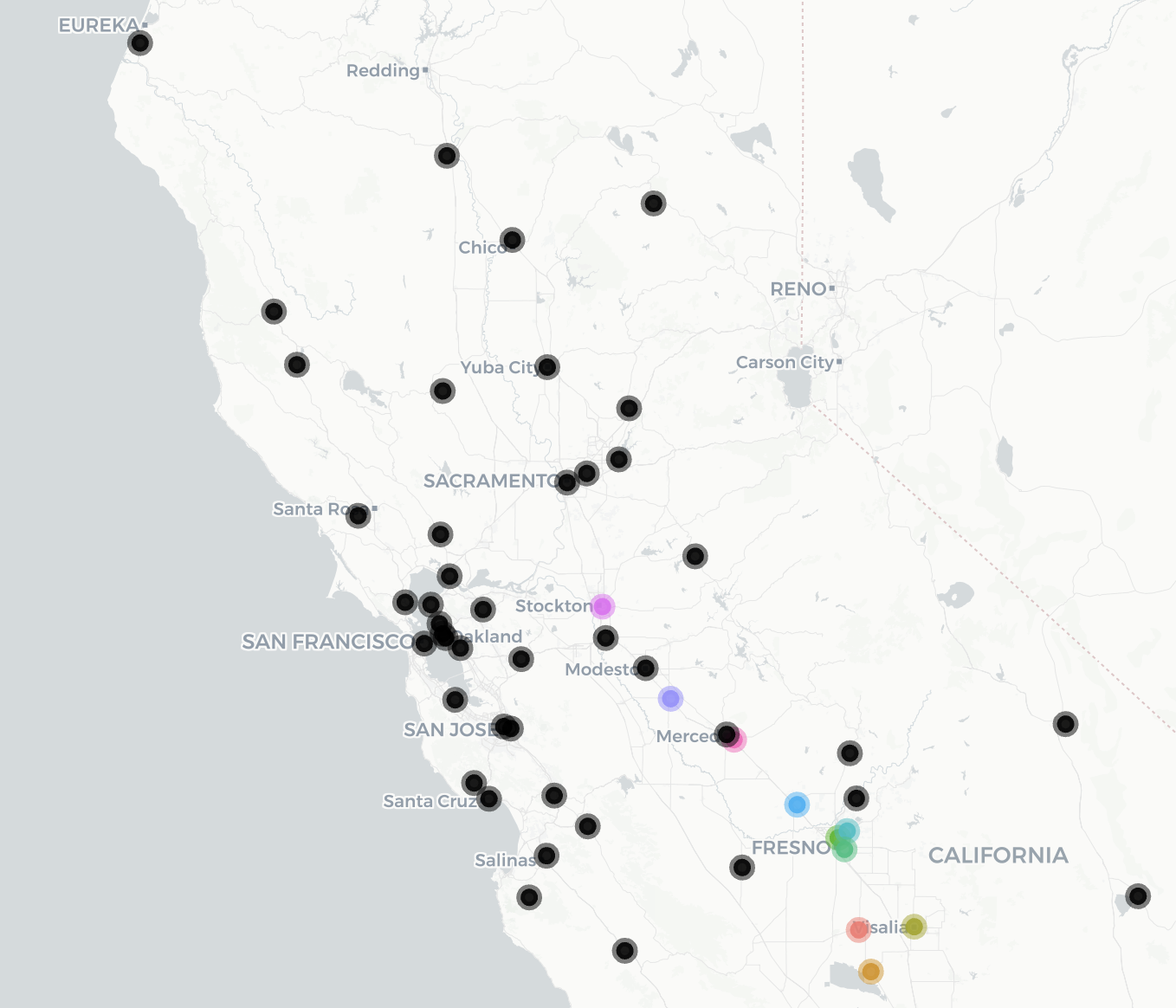}
    \caption{}
    \label{fig:sub1}
  \end{subfigure}\hfill
  \begin{subfigure}[t]{0.55\textwidth}
    \centering
    \includegraphics[width=\linewidth]{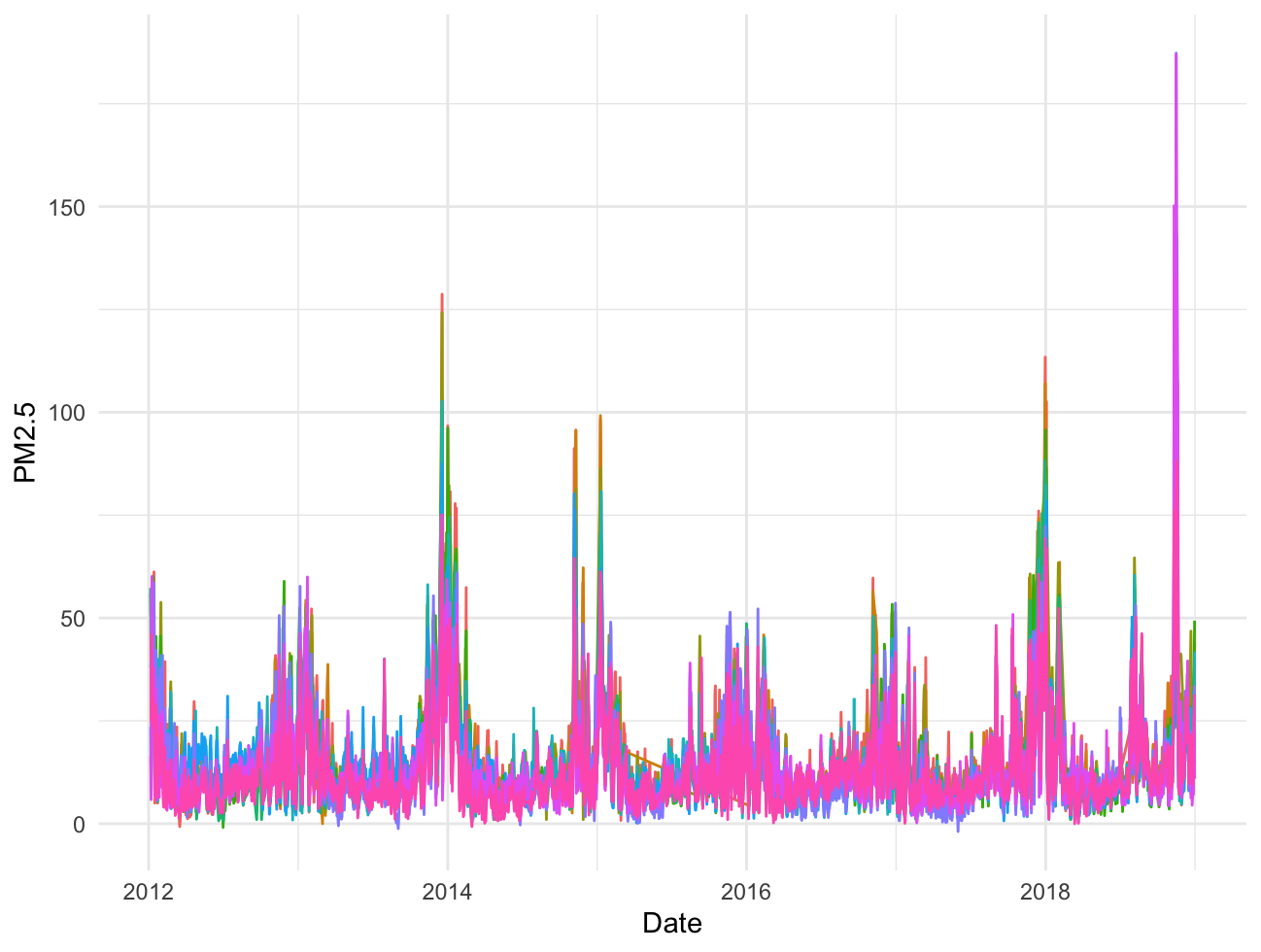}
    \caption{}
    \label{fig:sub2}
  \end{subfigure}
  \label{fig:both}
  \caption{Description of the PM$_{2.5}$ data (a) Locations of the 53 sites in Northern California used in this study, colored dots match (b), time series of the daily concentrations for the 10 monitoring sites with highest levels across 2012-2018.}

\end{figure}

Daily PM$_{2.5}$ measurements from 2012-2018, collected by regulatory monitors in California, were acquired from the EPA's Air Quality System (AQS) (\cite{AirData}). Where multiple instruments measured  PM$_{2.5}$ at one site (distinguished by Parameter Occurrence Code, POC), we averaged across POC so that each site had one PM$_{2.5}$ value per day.\\
There were 96 sites statewide from 2012-2018, and we retained 53 in northern California by filtering to latitude $>$ 35.75 (Figure \ref{fig:sub1}). There is a clear spike in PM$_{2.5}$ during the 2018 wildfire season, with smaller spikes in other years (Figure \ref{fig:sub2}). The abundance of outliers motivates using the Laplace kernel within the convolution smoothing method.

\subsubsection{Covariates}
We included 21 covariates capturing temporal, meteorological, and geographic variation. From AQS we use site longitude/latitude, the calendar day, month, and year, the Julian day, and 1–5 day lags of PM$_{2.5}$. Meteorological data for winds (direction, speed) were acquired from gridMET, which provides daily surface meteorological data for the contiguous United States at $\sim$4 km gridded resolution (\cite{Abatzoglou_2011}). Additional meteorology including precipitation, shortwave surface radiation, and mean temperature were available at a higher spatial resolution ($\sim$1 km) from Daymet \cite{thornton_daymet_2022}. Population counts were extracted from LandScan Global, which uses remote sensing-based global data modeling and mapping to create estimate at $\sim$1 km resolution (\cite{LandScan}). Also included were land cover \citep{friedl2010} and enhanced vegetation index (EVI) \citep{huete2002} from MODIS, and elevation from the Advanced Spaceborne Thermal Emission and Reflection Radiometer Global Digital Elevation Model (ASTER GDEM), which provides a digital elevation model for land at $\sim$30 m resolution \citep{Aster}.

\subsection{Model Tuning and Evaluation}
As in the simulation study, we used cross-validation to tune the hyperparameters using XGBoost with RMSE loss. We randomly split the data into training and testing sets, where 20\% of the data is used as the testing set and the remaining 80\% of the data is used as the training set. We then randomly sampled 100 hyperparameter combinations from a hyperparameter grid and ran 10-fold cross-validation (CV) on the training data to choose the best hyperparameters that produce the lowest RMSE. We then fix those values across all interval methods for comparability. 

To rigorously tune the smoothing hyperparameters, we utilized a spatial 5-fold cross-validation framework across the 53 monitoring sites. Similarly to the simulation study, we used the tuning grid \\ $\{0.10, 0.25, 0.50, 0.75, 1.0\}$ for the arctan and convolution methods. However, for the Huber method, we allowed for a wider tuning grid so that larger smoothing parameters could be tested (since the Huber method sometimes performs better with more smoothing). For the exceedance probabilities estimation, optimal parameters were selected to minimize the BSS, and we used the same penalty used in the simulation study to heavily penalize quantile crossings during tuning. We used both the Laplace and Gaussian kernels for the convolution smoothing method. The Python \texttt{xgboost} quantile implementation does not have any smoothing parameters for controlling coverage. 

As in the simulation study, we used LOSO-CV for all out-of-sample evaluations, this time across the 53 monitoring sites; the metrics reported represent the average performance across all 53 site-specific models. We evaluated both the multi-output tree method and the independent tree method for the dense probabilistic forecasts (predicting 9 discrete quantiles), reporting the average CRPS and the Quantile Crossing Rate for each. For the exceedance probabilities, derived from the same fitted quantiles, we report results under the multi-output tree method: the total number of observed and predicted exceedances ($N_{obs}$ and $N_{pred}$), the BSS, and the ROC-AUC score. Finally, as discussed in Section~\ref{sec:target-standardization}, target standardization (Z-scaling) was applied to stabilize the convolution kernels against very large values at the gradients when predicting at sites with very large $PM_{2.5}$ values. Because 2018 exhibited much larger variability, we only evaluated the 2018 dataset to test how our methods would perform with highly variable data.

\subsection{Results}
\subsubsection{Results for Dense CDF Estimation}
Table \ref{tab:real-crps-results} presents the results of the dense CDF analysis under the multi-output method, and Table \ref{tab:real-crps-results2} shows the independent tree method. The most striking result is the impact of the multi-output joint estimation on the average crossing rate. When estimating quantiles independently, severe quantile crossing occurs—up to 0.97 for the Huber pinball method and 0.84 for the convolution methods at lower smoothing parameters. By estimating the quantiles jointly with multi-output trees, these crossing rates drop significantly, reaching 0.00 for the arctan and convolution methods at moderate to high smoothing levels.

\begin{table}[H]
\caption{Dense CDF evaluation on the real PM$_{2.5}$ dataset using spatial LOSO-CV.}
\centering
\renewcommand{\arraystretch}{1.3}
\begin{tabularx}{\textwidth}{l >{\centering\arraybackslash}X *{2}{>{\centering\arraybackslash}X}}
\hline
\textbf{Method} & \textbf{Smoothing Parameter} & \textbf{Average CRPS} & \textbf{Average Crossing Rate} \\
\hline
Default Python         & \multirow{4}{*}{0.1} & 1.18 & 0.22 \\
Huber Pinball         &                      & 2.65 & 0.75 \\
Arctan Pinball         &                      & 1.20 & 0.02 \\
Convolution (Laplace)  &                      & 2.03 & 0.45 \\
Convolution (Gaussian) &                      & 2.53 & 0.57 \\
\hline
Default Python         & \multirow{4}{*}{0.5} & 1.18  & 0.22\\
Huber Pinball          &                      & 1.59 & 0.31 \\
Arctan Pinball         &                      & 1.79 & 0.00 \\
Convolution (Laplace)  &                      & 1.41 & 0.00 \\
Convolution (Gaussian) &                      & 1.36 & 0.00 \\
\hline
Default Python        & \multirow{4}{*}{1.0} & 1.18 & 0.22 \\
Huber Pinball         &                      & 1.21 & 0.04 \\
Arctan Pinball         &                      & 2.85 & 0.00 \\
Convolution (Laplace)  &                      & 1.99 & 0.00 \\
Convolution (Gaussian) &                      & 1.89 & 0.00 \\
\hline
\end{tabularx}
\label{tab:real-crps-results}
\end{table}

The smoothed methods also achieved lower Continuous Ranked Probability Scores (CRPS) while maintaining these zero crossing rates. For example, at a smoothing parameter of 0.5, the convolution methods scored CRPS values of 1.36 to 1.41, outperforming the Huber and Arctan pinball loss methods and providing sharper, better calibrated distribution estimates without the crossing issues of the independent models. However, it is essential to tune the smoothing parameter, since for the crossing rates for the convolution methods are quite large when the smoothing parameter is 0.1. At the lowest smoothing parameter tested (0.1), the Huber pinball loss had the largest average CRPS and crossing rate of all methods when trained independently. Its CRPS improved substantially at higher smoothing levels, though its crossing rate remained comparatively high relative to the arctan and convolution methods.

\subsection{Results for Exceedance Probabilities}
The exceedance probability results validate our ability to identify when values pass the $35.0~\mu g/m^{3}$ threshold. As shown in Table \ref{tab:exceedance-results}, Direct binary classification (Logistic baseline) overpredicts risk. Extracting probabilities using Gaussian kernel convolution smoothed XGB yields the best calibration and the highest Brier Skill Score. The convolution smoothed quantile XGB method also provides accurate exceedance probability estimates compared to the default Python quantile XGB method.

\begin{table}[H]
\caption{Exceedance Probability Results.}
\centering
\renewcommand{\arraystretch}{1.3}
\begin{tabularx}{\textwidth}{>{\raggedright\arraybackslash}X c c c c c}
\hline
\textbf{Model}                  & \textbf{Tuned Smoothing Parameter} & $\bm{N_{obs}}$ & $\bm{N_{pred}}$ & \textbf{BSS} & \textbf{ROC-AUC} \\
\hline
Logistic (Binary Baseline)      &       & 736 & 1343.90 & 0.41 & 0.99 \\
Method 1: QXGB (Default)        &       & 736 & 672.40  & 0.66 & 0.98 \\
Method 2: QXGB (Huber)          & t=1.0 & 736 & 643.40  & 0.69 & 0.98 \\
Method 3: QXGB (Arctan)         & s=0.1 & 736 & 977.40  & 0.67 & 0.98 \\
Method 4: QXGB (Gaussian Tuned) & h=0.5 & 736 & 747.20  & 0.70 & 0.99 \\
Method 5: QXGB (Laplace Tuned)  & h=0.2 & 736 & 687.90  & 0.69 & 0.99 \\
\hline
\end{tabularx}
\label{tab:exceedance-results}
\end{table}

\subsection{Results for Interval Estimation}
We assessed the standard 90\% intervals (estimating the 0.05 and 0.95 quantiles) for the wildfire data. For the intervals, we report PICP, PINAW, and the Interval Score. Table \ref{tab:app-results} presents the results of the 90\% intervals for 2018 under the multi-output spatial LOSO-CV, while Table \ref{tab:app-results2} in the Appendix presents the results using independent trees. 

The arctan pinball and convolution smoothing methods achieve PICP $\ge$ 0.9, while the Huber pinball and default Python methods fall short of the nominal 90\% target (PICP 0.74 and 0.87, respectively). Consistent with the sensitivity to smoothing parameters noted above, the arctan method's interval width becomes extreme in this out-of-sample spatial setting: its PINAW of 1.02 means the average interval exceeds the full observed range of the data, making it the least useful of the methods despite achieving perfect coverage. The default Python and Huber pinball methods achieve much lower PINAW and interval scores than the other methods, but this is driven in part by their undercoverage rather than genuinely tighter, well-calibrated intervals.

\begin{table}[H]
\caption{PICP and PINAW metrics for the intervals made by estimating the 0.05 and 0.95 quantiles using the 5 methods mentioned previously. This was done for the data from 2018.}
\centering
\renewcommand{\arraystretch}{1.3}
\begin{tabularx}{\textwidth}{p{1.8cm} X X X X X}
\hline
\textbf{Metric} & \textbf{Python} & \textbf{Huber \mbox{Pinball}} & \textbf{Arctan \mbox{Pinball}} & \textbf{\mbox{Convolution} \mbox{(Laplace)}} & \textbf{\mbox{Convolution} \mbox{(Gaussian)}} \\
\hline
\textit{PICP} & 0.87 & 0.74 & 1.00 & 0.97 & 0.98 \\
\textit{PINAW} & 0.14 & 0.10 & 1.02 & 0.24 & 0.31 \\
\textit{Interval Score} & 20.61 & 24.40 & 84.69 & 25.55 & 29.97 \\
\textit{Tuned Smoothing Parameter} &  & t=3.0 & s=0.1 & h=0.25 & h=0.5 \\
\hline
\end{tabularx}
\label{tab:app-results}
\end{table}

The following figures show the PM$_{2.5}$ levels in northern California in 2018, with lines showing the mean estimates of the upper and lower bounds of the intervals generated by the methods described previously. The intervals for all methods are able to capture the spikes in PM$_{2.5}$ values throughout the year. It is clear that the arctan method creates a prediction interval that is much too wide to be useful. The convolution methods create intervals that cover almost all the points, but that are more conservative and wider than the Huber and Arctan methods.

\begin{figure}[h!]
 \centering
  \begin{subfigure}[t]{0.5\textwidth}
    \centering
    \includegraphics[width=\linewidth]{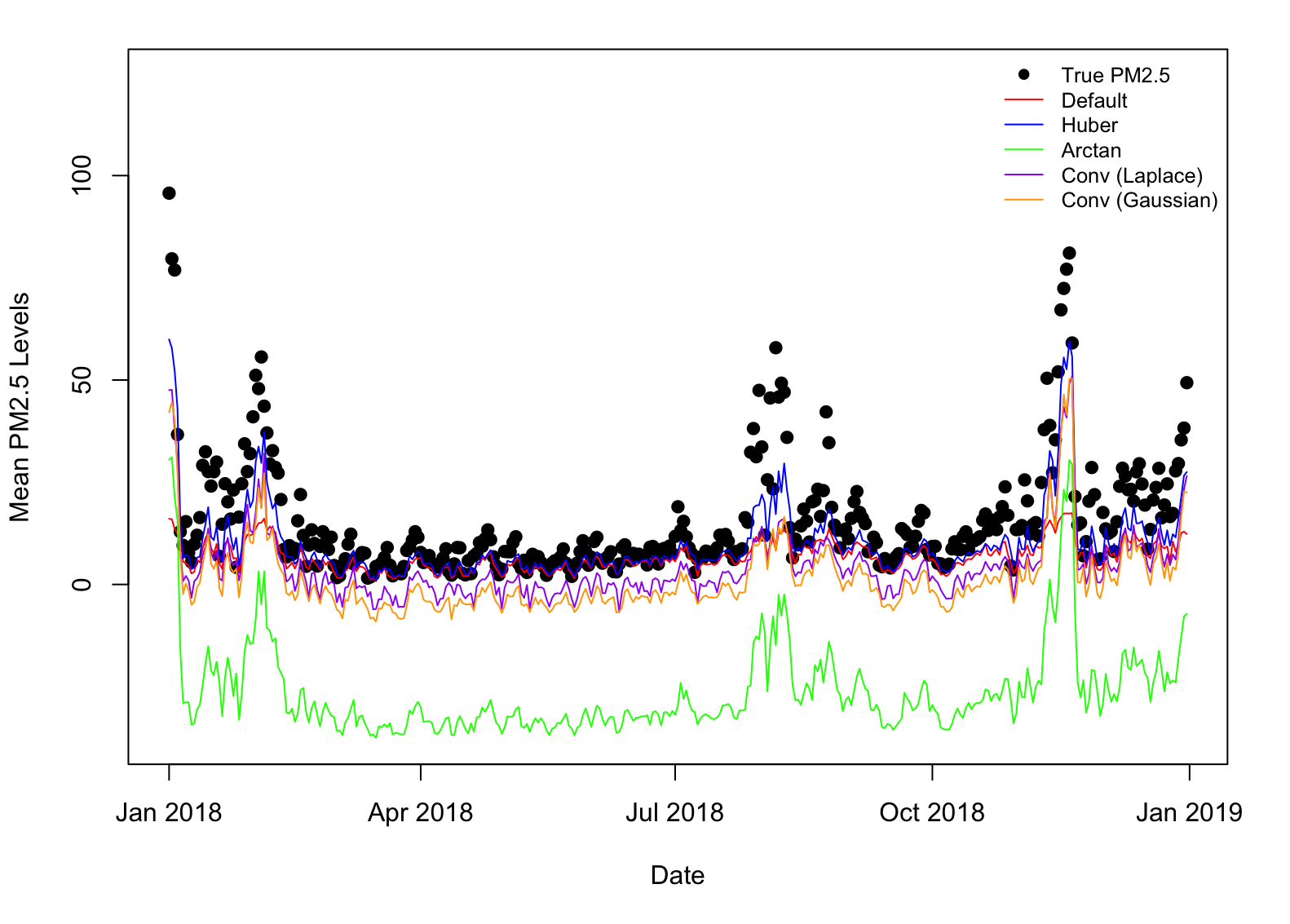}
    \caption{}
    \label{fig:real-lower}
  \end{subfigure}\hfill
  \begin{subfigure}[t]{0.5\textwidth}
    \centering
    \includegraphics[width=\linewidth]{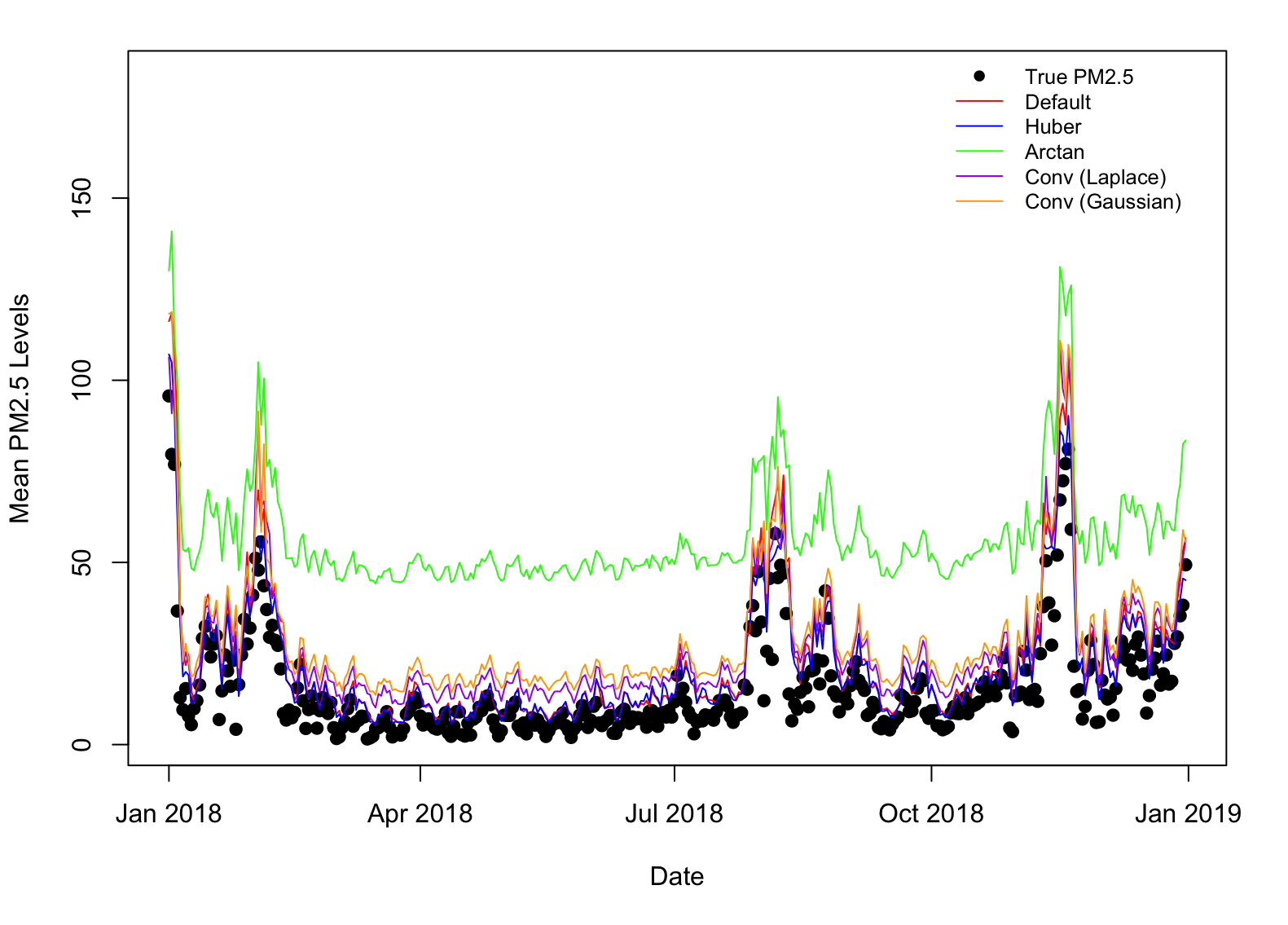}
    \caption{}
    \label{fig:real-upper}
  \end{subfigure}
  \label{fig:}
  \caption{Plots of 2018 PM$_{2.5}$ levels in Northern California, with lines showing the estimates of the (a) lower bounds of the intervals generated by the 5 approaches and (b) upper bounds of the intervals generated by the 5 approaches.}

\end{figure}

\section{Discussion}

\subsection{Methods Comparison}

Smoothed pinball approximations (Huber, arctan, and convolution) can be trained with the default XGBoost machinery, allowing a single boosted ensemble to estimate multiple quantiles separately or jointly. Sharing tree structures across quantiles improves computational efficiency and reduces quantile crossing because splits are coupled. While crossing between the 0.05 and 0.95 quantiles in the case of interval prediction is rare in practice because they are very separated, it can still occur locally in feature space, where joint estimation could mitigate this issue. Using second order information is important. XGBoost's split selection and Newton updates rely on both gradients and Hessians. Purely first order schemes (e.g. Python implementation of the quantile objective) forgo this and typically converge more slowly and choose less informative splits.
Conversely, too much smoothing can yield biased quantile estimates, while too little smoothing can yield noisy Hessians and unstable gradients.

The Python xgboost package and the smoothed approximation methods (Huber and arctan pinball loss, convolution smoothing) may perform better in different circumstances, and as noted above, too much smoothing of the loss function can result in biased quantile estimates. As shown by \cite{sluijterman2024}, estimating the lower and upper quantiles jointly with multi-output trees, rather than independent additive trees, can further reduce crossing; in this study we evaluate both tree strategies for every method. Our empirical results validate this: the multi-output method mostly produced near 0 crossing rates, even in out-of-distribution spatial test sets where independent trees had large crossing rates. It can also effectively regularize the models, as it can capture correlations between outputs (e.g. estimated quantile values), and it can increase computational efficiency. However, using separate models for each quantile can result in a better fit for each quantile. The multi-output trees would not perform well in cases where the outputs are very different from each other. However, in the case of estimating upper and lower quantiles, this is less likely to be an issue. The arctan pinball loss method \citep{sluijterman2024} and the convolution smoothing method we propose are the most suitable for using multi-output trees since the loss functions are smooth and have meaningful second derivatives everywhere. This helps to stabilize xgboost updates, which is needed as multiple outputs are being updated simultaneously.

\subsection{Convolution Method: Practical Guidance}

Kernel choice in convolution smoothing is data-dependent, which is an advantage of this method over others, since the kernel can be customized to match the data. Another finding is that since convolution smoothing relies on exponential functions, if an unscaled error is too large, it can trigger floating-point overflow and gradient instability, as evidenced by initial high crossing rates (20-45\%) on raw data. However, we demonstrated that this is easily remedied via target standardization. When the target data is standardized to a standard-normal range, convolution methods perform well, yielding CRPS scores highly competitive with Arctan while offering the flexibility to explicitly match the data's tails (e.g., Laplace kernel for heavy tails).

%\subsection{Convolution Method Bandwidth and Smoothing Choices}
Rule-of-thumb bandwidths such as those proposed by \cite{He_Pan_Tan_Zhou_2023} were a useful starting point for the convolution smoothing methods. To get more precisely tuned smoothing parameters, we performed empirical tuning using a spatial 5-fold cross-validation. In addition, integrating a penalty for quantile crossing into the tuning objective function discouraged smoothing parameter choices that could lead to quantile crossings.

%\subsection{Practical Guidance}
For interval estimation, in situations where coverage (PICP) falls short of the nominal level, we can increase the smoothing of the pinball loss to stabilize the Hessians, which in turn helps the achieved coverage more closely approach the nominal target. In situations where estimating a dense CDF is required, using multi-output trees rather than independent models is the most effective way to reduce quantile crossing. Furthermore, for convolution quantile regression xgboost methods, target standardization is recommended to prevent gradient instability resulting in high crossing rates. Finally, we recommend tuning smoothing parameters using spatial folds and heavily penalizing crossing in the evaluation metric.

\subsection{Limitations and Future Work}

One limitation is that it is more difficult to obtain accurate estimates for extreme values. This was evident in the exceedance probability results for the simulation study when the threshold was increased to $50.0~\mu g/m^{3}$. In that case, the results were much worse than the original results with the $35.0~\mu g/m^{3}$ threshold. This is likely due to the sparsity of the data in the far right tail of the distribution.

In the future, it may be useful to investigate asymmetric, mixture, or skew-adaptive kernels within the convolution smoothing method. For example, a skewed Laplace or a flexible asymmetric kernel with tunable tails could better fit the pronounced right-skew of extreme wildfire PM$_{2.5}$ events without sacrificing median accuracy.

We fixed the tree hyperparameters across methods for fairness and to isolate the effect of the loss function and smoothing on model performance. Holding these constant avoids confounding; otherwise, a method might look better simply because it was granted deeper trees or a milder regularizer, rather than because its quantile loss is better suited to the data. This mirrors our simulation design where we varied only the loss/smoothing while keeping the booster template identical.

Different smoothed losses can prefer slightly different booster settings. For example, smooth losses (arctan, convolution) often have a somewhat larger learning rate or fewer boosting rounds because second-order updates are more stable. 

Allowing per-method hyperparameter tuning could yield marginal gains. 

\newpage
\bibliographystyle{apalike}
\bibliography{references}

\newpage
\appendix

\section{Appendix A}
\setcounter{table}{0}
\renewcommand{\thetable}{A\arabic{table}}
\subsection{Other Kernels}
\begin{enumerate}
    \item Logistic kernel \(K(u) = \frac{e^{-u}}{(1 + e^{-u})^2}\):
  \[
    \ell_h(u) = \frac{h}{2}\,\ell^L(u/h) + \Bigl(\tau - \tfrac{1}{2}\Bigr)u,
    \quad \ell^L(u) := u + 2\log(1 + e^{-u}).
  \]
  \[
\ell_h'(u) = \tau - \frac{1}{1 + e^{u/h}}, \quad
\ell_h''(u) = \frac{1}{h} \cdot \frac{e^{u/h}}{(1 + e^{u/h})^2}
\]
  \item Uniform kernel \(K(u) = \tfrac{1}{2}\mathbf{1}(|u|\le1)\):
  \[
    \ell_h(u) = \frac{h}{2}\,\ell^U(u/h) + \Bigl(\tau - \tfrac{1}{2}\Bigr)u,
    \quad \ell^U(u) := \bigl(\tfrac{u^2}{2} + \tfrac{1}{2}\bigr)\mathbf{1}(|u|\le1) + |u|\mathbf{1}(|u|>1).
  \]
\[\ell_h'(u) =
\begin{cases}
\tau & u > h \\
\tau - \frac{1 - u/h}{2} & -h \leq u \leq h \\
\tau - 1 & u < -h
\end{cases},
\quad
\ell_h''(u) =
\begin{cases}
\frac{1}{2h} & |u| \leq h \\
0 & \text{otherwise}
\end{cases}\]
  \item Epanechnikov kernel \(K(u) = \frac{3}{4}(1 - u^2)\mathbf{1}(|u|\le1)\):
  \[
    \ell_h(u) = \frac{h}{2}\,\ell^E(u/h) + \Bigl(\tau - \tfrac{1}{2}\Bigr)u,
    \quad \ell^E(u) := \Bigl(\frac{3u^2}{4} - \frac{u^4}{8} + \frac{3}{8}\Bigr)\mathbf{1}(|u|\le1) + |u|\mathbf{1}(|u|>1).
  \]
  \[
\ell_h'(u) =
\begin{cases}
\tau & u > h \\
\tau - \left[ \frac{3}{4} \left( -\frac{u}{h} - \frac{(-u/h)^3}{3} \right) + \frac{1}{2} \right] & |u| \leq h \\
\tau - 1 & u < -h
\end{cases}
\]
\[
\ell_h''(u) =
\begin{cases}
\frac{3}{4h} \left(1 - \left( \frac{u}{h} \right)^2 \right) & |u| \leq h \\
0 & \text{otherwise}
\end{cases}
\]
  \item Triangular kernel \(K(u) = (1 - |u|)\mathbf{1}(|u|\le1)\):
  \[
    \ell_h(u) = \frac{h}{2}\,\ell^T(u/h) + \Bigl(\tau - \tfrac{1}{2}\Bigr)u,
    \quad \ell^T(u) := \Bigl(u^2 - \frac{|u|^3}{3} + \frac{1}{3}\Bigr)\mathbf{1}(|u|\le1) + |u|\mathbf{1}(|u|>1).
  \]
  \[
\ell_h'(u) =
\begin{cases}
\tau & u > h \\
\tau - \frac{(-u/h + 1)^2}{2} & -h \leq u < 0 \\
\tau - \left(1 - \frac{(1 + u/h)^2}{2} \right) & 0 \leq u \leq h \\
\tau - 1 & u < -h
\end{cases}
\]
\[
\ell_h''(u) =
\begin{cases}
\frac{1}{h} \left(1 - \left| \frac{u}{h} \right| \right) & |u| \leq h \\
0 & \text{otherwise}
\end{cases}
\]

\end{enumerate}

It is interesting to note that the Uniform kernel yields a smoothed loss function that is similar to the Huber pinball loss function.

\subsection{Sensitivity Analysis for Simulation Study Intervals}
\begin{table}[H]
\centering
\caption{Combined performance metrics with varying smoothing parameters and intervals}
\label{tab:sim-interval-sensitivity}
\begin{tabular}{l S[table-format=4.3] S[table-format=1.4] S[table-format=1.4] S[table-format=3.3]}
\hline
\textbf{Method} & \textbf{Smoothing Parameter} & \textbf{PICP} & \textbf{PINAW} & \textbf{Interval Score} \\
\hline
Huber 
    & 0.1 & 0.82 & 0.18 & 57.75 \\
    & 0.25  & 0.77 & 0.16 & 52.85  \\
    & 0.5  & 0.73 & 0.14 & 55.71  \\
    & 0.75   & 0.70 & 0.14 & 56.81  \\
    & 1.0   & 0.69 & 0.13 & 57.69  \\
\hline
Arctan pinball loss 
    & 0.1 & 0.97 & 0.25 & 53.86  \\
    & 0.25   & 0.99 & 0.42 & 75.49  \\
    & 0.5  & 0.99 & 0.75 & 121.16  \\
    & 0.75    & 1.00 & 1.09 & 169.56  \\
    & 1.0   & 1.00 & 1.43 & 219.05  \\
\hline
Convolution (Laplace) 
    & 0.1 & 0.91 & 0.19 & 50.26  \\
    & 0.25  & 0.96 & 0.22 & 52.40  \\
    & 0.5  & 0.98 & 0.32 & 63.26  \\
    & 0.75    & 0.99 & 0.43 & 77.89  \\
    & 1.0   & 0.99 & 0.55 & 93.98  \\
\hline
Convolution (Gaussian)
    & 0.1 & 0.90 & 0.18 & 52.00  \\
    & 0.25 & 0.94 & 0.20 & 51.20  \\
    & 0.5 & 0.97 & 0.26 & 56.79  \\
    & 0.75 & 0.98 & 0.34 & 65.58  \\
    & 1.0 & 0.99 & 0.42 & 75.96  \\
\hline
\end{tabular}
\end{table}

\subsection{Simulation Study Results for Exceedance Probabilities, Threshold=50.0}

\begin{table}[H]
\caption{Exceedance Probability Results for Threshold=50.0}
\centering
\renewcommand{\arraystretch}{1.3}
\begin{tabularx}{\textwidth}{>{\raggedright\arraybackslash}X c c c c c}
\hline
\textbf{Model}                  & \textbf{Tuned Smoothing Parameter} & $\bm{N_{obs}}$ & $\bm{N_{pred}}$ & \textbf{BSS} & \textbf{ROC-AUC} \\
\hline
Logistic (Binary Baseline)      &       & 386 & 2114.90 & -0.84 & 0.81 \\
Method 1: QXGB (Default)        &       & 386 & 1067.70  & 0.25 & 0.75 \\
Method 2: QXGB (Huber)          & t=0.5 & 386 & 1064.80  & 0.24 & 0.73 \\
Method 3: QXGB (Arctan)         & s=0.2 & 386 & 1114.90  & 0.24 & 0.79 \\
Method 4: QXGB (Gaussian Tuned) & h=0.05 & 386 & 1059.80  & 0.24 & 0.74 \\
Method 5: QXGB (Laplace Tuned)  & h=0.5 & 386 & 1089.60  & 0.25 & 0.76 \\
\hline
\end{tabularx}
\label{tab:exceedance-results-sim50}
\end{table}

\subsection{Individual Tree Method Results}
\begin{table}[H]
\caption{PICP, PINAW, and interval score for 90\% prediction intervals formed from the 0.05 and 0.95 quantiles, quantiles computed individually}
\centering
\renewcommand{\arraystretch}{1.3}
\begin{tabularx}{\textwidth}{p{1.7cm} *{5}{>{\centering\arraybackslash}X}}
\hline
\textbf{Simulated Data} & \textbf{Python} &\textbf{Huber Pinball} & \textbf{Arctan Pinball} & \textbf{Convolution (Laplace)} & \textbf{Convolution (Gaussian)} \\
\hline
\multicolumn{6}{>{\raggedright\arraybackslash}p{\linewidth}}{\textbf{With Spike}}\\
\textit{PICP} & 0.86  & 0.75 & 0.96 & 0.95 & 0.93 \\
\textit{PINAW} & 0.19  & 0.15 & 0.24 & 0.22 & 0.20 \\
\textit{Interval Score} & 50.42 & 57.48 & 53.36 & 52.80 & 51.02 \\
\textit{Tuned Smoothing Parameter} &  & t=0.25 & s=0.1 & h=0.25 & h=0.25 \\
\hline
\end{tabularx}
\label{tab:sim-results2}
\end{table}

\begin{table}[H]
\caption{Average CRPS and Crossing Rates across all 50 simulated locations, quantiles computed individually}
\centering
\renewcommand{\arraystretch}{1.3}
\begin{tabularx}{\textwidth}{l >{\centering\arraybackslash}X *{2}{>{\centering\arraybackslash}X}}
\hline
\textbf{Method} & \textbf{Smoothing Parameter} & \textbf{Average CRPS} & \textbf{Average Crossing Rate} \\
\hline
Default Python         & \multirow{4}{*}{0.1} & 2.71 & 0.13 \\
Huber Pinball         &                      & 2.79 & 0.69 \\
Arctan Pinball         &                      & 2.74 & 0.19 \\
Convolution (Laplace)  &                      & 2.74 & 0.36 \\
Convolution (Gaussian) &                      & 2.75 & 0.43 \\
\hline
Default Python         & \multirow{4}{*}{0.5} & 2.71    & 0.13\\
Huber Pinball          &                      & 2.79 & 0.28 \\
Arctan Pinball         &                      & 3.22 & 0.01 \\
Convolution (Laplace)  &                      & 2.83 & 0.02 \\
Convolution (Gaussian) &                      & 2.77 & 0.02 \\
\hline
Default Python        & \multirow{4}{*}{1.0} & 2.71 & 0.13 \\
Huber Pinball         &                      & 2.81 & 0.14 \\
Arctan Pinball         &                      & 4.34 & 0.00 \\
Convolution (Laplace)  &                      & 3.33 & 0.00 \\
Convolution (Gaussian) &                      & 3.17 & 0.00 \\
\hline
\end{tabularx}
\label{tab:sim-crps-results2-ind}
\end{table}

\begin{table}[H]
\caption{PICP and PINAW metrics for the prediction intervals made by estimating the 0.05 and 0.95 quantiles using the 5 methods mentioned previously. This was done for the data from 2018, quantiles computed individually}
\centering
\renewcommand{\arraystretch}{1.3}
\begin{tabularx}{\textwidth}{p{1.8cm} X X X X X}
\hline
\textbf{Metric} & \textbf{Python} & \textbf{Huber \mbox{Pinball}} & \textbf{Arctan \mbox{Pinball}} & \textbf{\mbox{Convolution} \mbox{(Laplace)}} & \textbf{\mbox{Convolution} \mbox{(Gaussian)}} \\
\hline
\textit{PICP} & 0.91 & 0.74 & 0.99 & 0.97 & 0.98 \\
\textit{PINAW} & 0.26 & 0.11 & 0.64 & 0.24 & 0.30 \\
\textit{Interval Score} & 32.56 & 25.89 & 66.11 & 25.39 & 30.00 \\
\textit{Tuned Smoothing Parameter} &  & t=10.0 & s=0.1 & h=0.25 & h=0.5 \\
\hline
\end{tabularx}
\label{tab:app-results2}
\end{table}

\begin{table}[H]
\caption{Dense CDF evaluation on the real PM$_{2.5}$ dataset using spatial LOSO-CV, quantiles computed individually}
\centering
\renewcommand{\arraystretch}{1.3}
\begin{tabularx}{\textwidth}{l >{\centering\arraybackslash}X *{2}{>{\centering\arraybackslash}X}}
\hline
\textbf{Method} & \textbf{Smoothing Parameter} & \textbf{Average CRPS} & \textbf{Average Crossing Rate} \\
\hline
Default Python         & \multirow{4}{*}{0.1} & 1.33 & 0.36 \\
Huber Pinball         &                      & 2.60 & 0.97 \\
Arctan Pinball         &                      & 1.20 & 0.29 \\
Convolution (Laplace)  &                      & 1.82 & 0.75 \\
Convolution (Gaussian) &                      & 2.32 & 0.84 \\
\hline
Default Python         & \multirow{4}{*}{0.5} & 1.33  & 0.36\\
Huber Pinball          &                      & 1.53 & 0.82 \\
Arctan Pinball         &                      & 1.80 & 0.02 \\
Convolution (Laplace)  &                      & 1.42 & 0.03 \\
Convolution (Gaussian) &                      & 1.37 & 0.05 \\
\hline
Default Python        & \multirow{4}{*}{1.0} & 1.33 & 0.36 \\
Huber Pinball         &                      & 1.22 & 0.36 \\
Arctan Pinball         &                      & 2.85 & 0.01 \\
Convolution (Laplace)  &                      & 1.99 & 0.01 \\
Convolution (Gaussian) &                      & 1.89 & 0.01 \\
\hline
\end{tabularx}
\label{tab:real-crps-results2}
\end{table}

\section{Appendix B}
\subsection{Gradient Boosting}
Given a dataset with $n$ samples: $(x_1, y_1), (x_2, y_2), ..., (x_n, y_n)$,
gradient boosting builds an ensemble of decision trees in an additive manner, with the aim of finding a function $F(x)$ that minimizes the loss function $L(y, F(x))$ (\cite{Friedman_2001}). We initialize the model $F_0(x)$ by finding the constant that minimizes the loss function:

\[F_0(x) = \arg\min_c \sum_{i=1}^{n} L(y_i, c).\]

For a standard mean squared error loss function, $F_0(x)$ is simply the sample mean. Then, we repeat the following process for each iteration $m$:

\begin{enumerate}
    \item The (pseudo-)residuals for the current model are computed, so that at $F_{m-1}(x)$, we get the best steepest-descent step direction by using the negative gradient of the loss function:
    \[r_i^{(m)} = - \left[ \frac{\partial L(y_i, F(x_i))}{\partial F(x_i)} \right]_{F(x) = F_{m-1}(x)}.\]
    For the mean squared error loss function, $r_i^{(m)} = y_i - F_{m-1}(x_i)$.
    \item We fit a new decision tree $T_m(x)$ to the (pseudo-)residuals so that it minimizes 
    \[T_m = \arg\min_T \sum_{i=1}^{n} \left( r_i^{(m)} - T(x_i) \right)^2.\]
    \item We find the optimal learning rate 
    \[\eta_m=\arg\min_{\eta} \sum_{i=1}^{n} L(y_i, F_{m-1}(x_i) + \eta T_m(x_i)), \: 0<\eta_m\leq1,\]
    which controls how much the new tree contributes to the current model.
    \item  We update the new model by adding the weighted new tree to the current model:
    \[F_m(x) = F_{m-1}(x) + \eta_m T_m(x)\]

It is crucial to include the learning rate $\eta$ to prevent overfitting.

\end{enumerate}

\subsection{Extreme Gradient Boosting}
The XGBoost algorithm is a special version of gradient boosting that is optimized, faster, and more accurate. For example, it supports parallel processing and uses hardware optimization (e.g. GPU acceleration) for faster training, and it handles missing values effectively (\cite{Chen_Guestrin_2016}). We highlight two important changes to the algorithm that reduce overfitting and improve optimization:

\subsubsection{Regularization}
Instead of having the loss function as the objective function, we now have an objective function that consists of both a loss function and a regularization term. The addition of the regularization term ensures that we do not overfit by having too many overly complex trees. The new objective function is given by

\[L_m = \sum_{i=1}^{n} L(y_i, F_m(x_i)) + \Omega(T_m),\]

with the regularization term

\[\Omega(T_m) = \gamma N + \frac{1}{2} \lambda \sum_j w_j^2,\]

where $\gamma$ is a regularization parameter that controls the complexity of the trees, $N$ is the number of leaves in the tree, and $\lambda$ is a parameter that penalizes the squared weight of the leaves $w_j$. These weights are obtained in closed form from the first- and second-order terms (gradients $g_i$ and Hessians $h_i$) in the Taylor expansion used by XGBoost.

\subsubsection{Second-Order Gradient}
Instead of using only the first-order gradient of the loss function with the pseudo-residuals in gradient boosting, XGBoost uses both the first-order gradient $g_i = \frac{\partial L(y_i, F_{m-1}(x_i))}{\partial F_{m-1}(x_i)}$ and the second-order gradient (Hessian) $h_i = \frac{\partial^2 L(y_i, F_{m-1}(x_i))}{\partial(F_{m-1}(x_i))^2}$ by using a Taylor expansion so that instead of minimizing 

\[L_m = \sum_{i=1}^{n} L(y_i, F_m(x_i)) + \Omega(T_m),\]

we minimize 

\[
L_m\approx \sum_{i=1}^{n} 
\Bigl[\,L(y_i,F_{m-1}(x)) + 
g_i\,T_m(x_i)+\tfrac{1}{2}h_i\,T_m(x_i)^2\,\Bigr]+\Omega(T_m),\]

This can be simplified by removing the constant terms so that we get

\[\tilde{L}_m\approx \sum_{i=1}^{n} [g_iT_m(x_i)+\frac{1}{2}h_iT_m^2(x_i)]+\Omega(T_m).\]

The hessians are also used to compute the weights of each leaf $j$ in the regularization term,

\[w_j=-\frac{\sum_{i \in I_j} g_i}{\sum_{i \in I_j} h_i+\lambda}.\]

This can be used to compute the loss reduction for a given split using

\[L_{split}=\frac{1}{2}\left[\frac{(\sum_{i\in I_L} g_i)^2}{\sum_{i\in I_L} h_i+\lambda}+\frac{(\sum_{i\in I_R} g_i)^2}{\sum_{i\in I_R} h_i+\lambda}-\frac{(\sum_{i\in I} g_i)^2}{\sum_{i\in I} h_i+\lambda}\right]-\gamma,\]

where $I_L$ and $I_R$ are the instance sets of left and right nodes after the split, and $I=I_L\cup I_R$. This loss reduction is used to evaluate the split candidates.

\subsection{Quantile Regression}
Quantile regression is analogous to ordinary least squares, but it estimates conditional quantiles instead of estimating the conditional mean \citep{koenker1978}. It aims to minimize the pinball loss function 

\[
L_\tau(y,\hat{y}) = 
\begin{cases}
    \tau(y - \hat{y}), & y \geq \hat{y}\\
    (\tau-1)(y-\hat{y}), & y<\hat{y},
\end{cases}
\]

where $y$ is the observed value, $\hat{y}$ is the predicted value, and $\tau$ is the quantile of interest to estimate. 

\end{document}